\documentclass{article}
\usepackage{spconf,amsmath,graphicx}

\usepackage{url}
\usepackage{subfig}
\usepackage{amssymb}
\usepackage{array}
\usepackage{multirow}
\DeclareMathSizes{10}{10}{5}{5} 
\title{Multi-View Fusion and Distillation for Subgrade Distresses Detection based on 3D-GPR}
\name{Chunpeng Zhou $^{\star}$ \thanks{This research was supported by Zhejiang Provincial Key Research and Development Program of China under Grant No. 2021C01106.} 
\qquad Kangjie Ning $^{\star}$
\qquad Haishuai Wang $^{\dag}$ 
\qquad Zhi Yu $^{\dag}$
\qquad Sheng Zhou
\qquad Jiajun Bu
\thanks{$^{\star}$ These authors contributed equally.}
\thanks{$^{\dag}$ Corresponding Authors}
}

\address{Zhejiang University, China}
\begin{document}
%\ninept
%
\maketitle
\begin{abstract}
The application of 3D ground-penetrating radar (3D-GPR) for subgrade distress detection has gained widespread popularity. To enhance the efficiency and accuracy of detection, pioneering studies have attempted to adopt automatic detection techniques, particularly deep learning. However, existing works typically rely on traditional 1D A-scan, 2D B-scan or 3D C-scan data of the GPR, resulting in either insufficient spatial information or high computational complexity.  To address these challenges,  we introduce a novel methodology for the subgrade distress detection task by leveraging the multi-view information from 3D-GPR data. Moreover, we construct a real multi-view image dataset derived from the original 3D-GPR data for the detection task,  which provides richer spatial information compared to A-scan and B-scan data, while reducing computational complexity compared to C-scan data. Subsequently, we develop  a novel \textbf{M}ulti-\textbf{V}iew \textbf{V}usion and \textbf{D}istillation framework, \textbf{GPR-MVFD}, specifically designed to optimally utilize the multi-view GPR dataset. This framework ingeniously incorporates multi-view distillation and attention-based fusion to facilitate significant feature extraction for subgrade distresses. In addition, a self-adaptive learning mechanism is adopted to stabilize the model training and prevent performance degeneration in each branch. Extensive experiments conducted on this new GPR benchmark demonstrate the effectiveness and efficiency of our proposed framework. Our framework outperforms not only the existing GPR baselines, but also the state-of-the-art methods in the fields of multi-view learning, multi-modal learning, and knowledge distillation. We will release the constructed multi-view GPR dataset with expert-annotated labels and the source codes of the proposed framework. 
\end{abstract}
\section{Introduction}
\label{sec:intro}
Road transport infrastructure plays a crucial component in modern society. However, with the increase of service life period and the impact of the natural environmental factors, the associated safety risks and maintenance costs caused by road damage become increasingly significant and cannot be overlooked. Consequently, the 3-Dimension ground penetrating radar (3D-GPR), as a prominent detection tool, is increasingly applied for the infrastructure health monitor, particularly in the road detection \cite{khudoyarov2020three,rasol2022gpr}, bridge safety inspection \cite{wang2010automatic,bachiri2018bridge,DINH2018292} and airport runway inspection \cite{li2022mv}. Notably, the non-destructive nature of 3D-GPR enables the effective detection of subgrade distresses without compromising the integrity of the infrastructure. Nonetheless, the analysis and interpretation of the collected 3D-GPR data are highly challenging and require human experts with extensive experience and domain knowledge in this area, causing inefficiencies and potential inaccuracies.  Consequently, the cost of the analysis and interpretation will be prohibitively expensive. To overcome these challenges, the research community has sought to leverage the recent automatic detection techniques to  analyze  the 3D-GPR data, particularly deep learning \cite{wang2010automatic, bachiri2018bridge, rasol2022gpr}.

% The interpretation of GPR data mostly still relies on experienced human experts. Manual interpreting, however, is subjective, time-consuming, and cost-prohibitive, so it is not appropriate for large amounts of GPR data.
%On the other hand, with the fast development of the deep learning techniques, the deep learning has achieved distinguished performances in various downstream tasks, compared with the traditional machine learning methods, and even can surpass the humans performance. So, recent researches have try to utilize the deep learning to analyze and predict the data from 3D-GPR
% Locate underground objects and sub-surface conditions
% Inspection and maintenance of infrastructure
% crack void  Interlayer Disengaging

\begin{figure}[t]
    \centering
    \includegraphics[width=6cm]{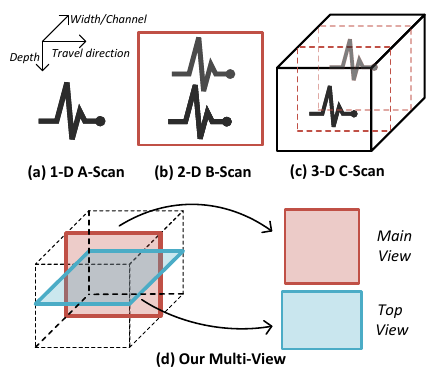}
    \caption{An illustration of different types of  3D-GPR data. An sample in our multi-view GPR dataset contains two views: the main view and the top view, which can be considered as the intermediate between the 2D B-scan data and 3D C-scan data.}
    \label{fig:1}
\end{figure}

\noindent\textbf{Related work: } Concurrently, the rapid development of deep learning techniques \cite{lecun2015deep,voulodimos2018deep} has enabled notable performance in various downstream tasks, outperforming traditional machine learning methods and even surpassing human performance in some cases \cite{He_2016_CVPR,Lin_2017_ICCV,He_2017_ICCV,bachiri2018bridge}. Accordingly, recent researches have aimed to employ deep learning to automatically analyse and predict the 3D-GPR data \cite{khudoyarov2020three,rasol2022gpr, tong2020advances,7572995,9702139}, which can significantly enhance maintenance efficiency while reducing detection costs.

% According the different types of the processed 3D-GPR data, the deep learning for 3D-GPR can be divided three kinds, including 
% 3-Dimension ground penetrating radar (3D-GPR)
% Our dataset can be seemed in between the 2D B-scan data and 3D C-scan data.

According to the different types of the processing approaches, the 3D-GPR equipment can generated different forms of GPR data, including the one-dimensional, two-dimensional and three-dimensional form of the echo signal to carry out the display of the vanilla 3D-GPR data, named as A-scan data, B-scan data, C-scan data respectively \cite{tong2020advances}. As depicted in Fig \ref{fig:1}, the A-scan data is 1D amplitude-time echo signal which records in a fixed location. Though some works attempt to exploit A-scan data with deep learning \cite{giannakis2019machine,tong2020pavement,wang2019human,rs13122375}, but the performance can not be satisfied due to the very limited information in a 1-D A-scan data. 
Therefore, most recent works shifted their focus towards the utilization of  the 2D B-scan data \cite{TONG201869,DINH2018292,su2023end}. When a 3D-GPR traverses the measurement line along the detection direction, it collected a series of A-scan data. After that, these collected A-scan data are spliced together to form the B-scan data. 
% When the 3D-GPR collects data from multiple measurement lines simultaneously, and the data obtained from each measurement line corresponds to a B-scan data. 
Even though the B-scan data contains more information compared to the A-scan data, it still possesses certain limitations as demonstrated in previous studies \cite{kim2021novel,li2022mv} . For instance, the 
 crack and void usually exhibit similar reflection patterns in B-scan images \cite{li2022mv}, Thus, it is difficult to distinguish them using only B-scan images.
Further, the C-scan data can be generated by juxtaposing of multiple B-scan data \cite{TONG201869, rs14071593}, which can also be viewed as the spatial combination of several B-scan data.  Thereby  the 3D C-scan data contains more spatial information.  However,  the collection and procession of  C-scan data is more complex than the A-scan or B-scan data. Particularly when utilizing 3D convolutional neural networks to extract the information from C-scan data, the computational complexity is quiet  high \cite{khudoyarov2020three, 9702139}, which limits its practical application range.

\noindent\textbf{Present work: } To abbreviate the aforementioned limitations, we aim to leverage spatial information while without substantially increasing computational complexity.
To achieve this, we build a multi-view GPR dataset, which are sampled from the 3D C-scan data. As depicted in Fig \ref{fig:1},  given a segmentation of the C-scan data, we first take a width (or channel ) slice of a plane, referred to as Main View, which can be regarded as a B-scan image. Subsequently, we take a plane slice along the distance, referred to as Top View. Therefore, the main view captures  the information of the travel-depth direction, while the top view captures  the information of the travel-width direction. Our built multi-view GPR dataset can be regarded as a set of  2D images, which contains more spatial information compared to A-scan, B-scan data, and reduce data computational complexity compared to C-scan data. This novel multi-view GPR dataset serves as the foundation for a new benchmark designed for subgrade distress detection.

Next, we propose a novel Multi-View Fusion and Distillation framework, referred to as  GPR-MVFD, tailored specifically to leverage the multi-view GPR dataset.  The proposed GPR-MVFD contains two branches, each corresponding to a view and serving to  extract feature from their respective view. Specially, we deploy an attention based fusion module that effectively combines information from both views, enabling the learning of different fusion weights and obtaining more robust representation of GPR data. Additionally, the multi-view distillation module take the advantages of the knowledge distillation technique \cite{hinton2015distilling,zhang2018deep, qian2022switchable} among two branches and the fusion module, allowing them to teach each other and enhance the learning process.
Notablely, due to the potential performance gap between two branches caused by the different inputs, the branch with better performance may degenerate \cite{qian2022switchable}. To avoid these situations, we introduce a self-adaptive learning mechanism, which can adaptively decide  the branch to train or not. Consequently, it can prevent performance degradation by halting parameter updates when necessary.

Then, we evaluate our proposed GPR-MVFD on the new multi-view GPR benchmark, and compare it with the existing GPR baselines, multi-view, multi-modal, and knowledge distillation-based methods. The experimental results reveal that the proposed GPR-MVFD achieves the state-of-the-art performance with the relative low computational complexity.

To these ends, our contributions can be summarized as following:

(\romannumeral1) we introduce a novel methodology for the subgrade distresses detection task based on the 3D-GPR data, which leverages  the multi-view information of the 3D-GPR data, rather than the traditional A-scan, B-scan or C-scan data. To the best of our knowledge, we are the first to build and publicly release a multi-view GPR dataset with expert-annotated labels. Additionally, a new benchmark for GPR subgrade distresses detection are established based on this new GPR dataset.
(\romannumeral2) we develop a new multi-branch framework, tailored specifically for the new multi-view GPR dataset. This framework incorporates multi-view distillation and attention-based fusion to enhancing  the learning of significant representations across each view. A self-adaptive learning mechanism was deployed within this framework to stabilize the learning process and prevent performance degeneration.
(\romannumeral3) we conduct extensive experiments on the new GPR benchmark to demonstrate the efficacy of our proposed framework. Our results surpass not only the existing traditional GPR baselines, but also including the state-of-the-art multi-view, multi-modal, and knowledge distillation-based methods.

\section{Multi-View GPR Dataset and Benchmark}
\begin{figure}[t]
    \centering
    \includegraphics[width=8.3cm]{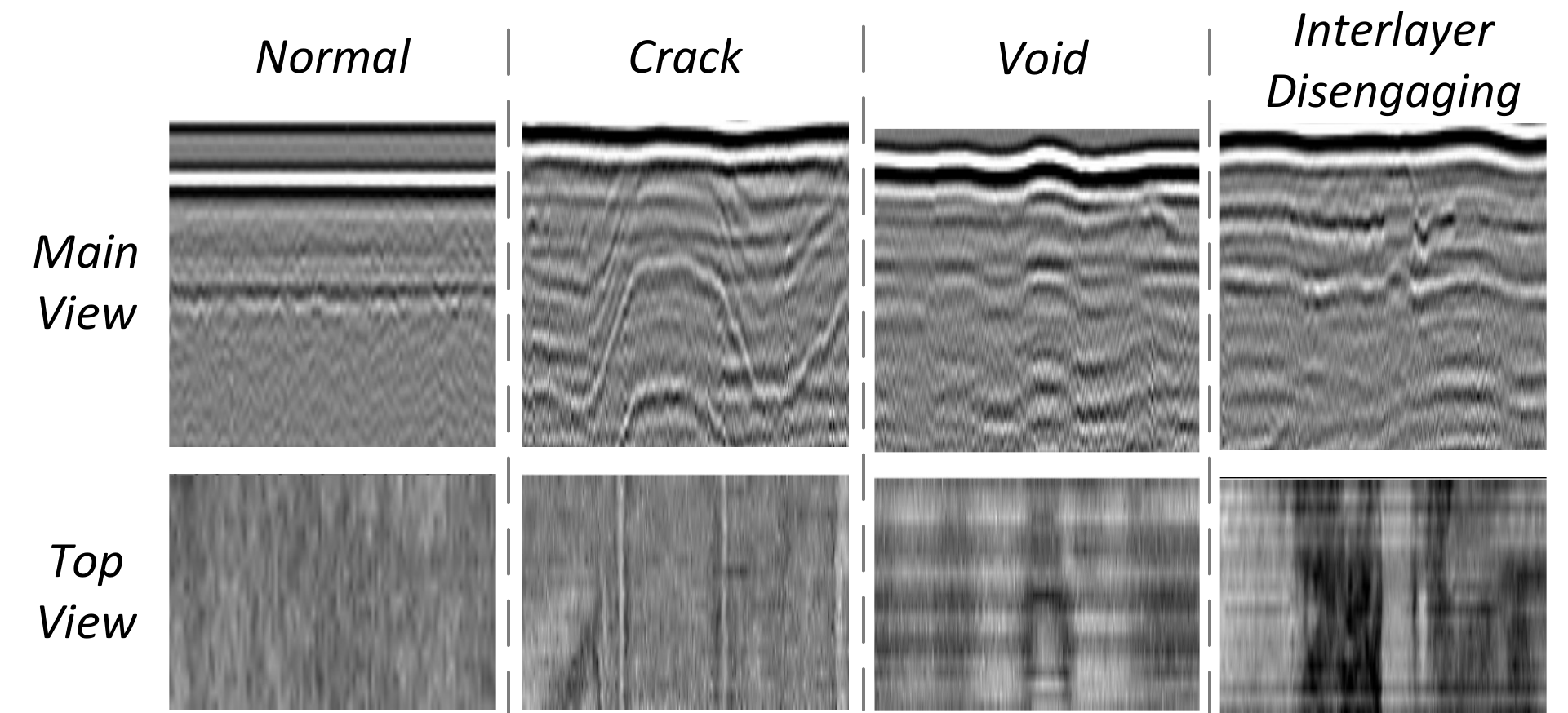}
    \caption{Examples of different subgrade distresses in our built multi-view GPR dataset. Each sample contains two view: the main view image and the top view image.}
    \label{fig:2}
\end{figure}

This real Multi-View 3D-GPR dataset was collected from urban roads in Zhejiang Province, China.  These data was gathered using a vehicle-mounted 3D-Radar ground-coupled antenna array\footnote{\url{http://3d-radar.com/}} with multichannel antenna. Subsequently, we produce multi-view data from the vanilla collected 3D-GPR data by using the 3D-Radar Examiner software\footnote{ \url{http://3d-radar.com/system/}}. Totally, we have collected  682 subgrade samples/segmentations in our Multi-View 3D-GPR dataset, where each sample contains two view images: the main-view image  with a resolution of 320 $\times$ 320 pixels, and the top-view image  with a resolution of 320 $\times$ 230 pixels. For the task of  subgrade distress detection, in addition to the normal samples without distress, we aim to identify three types of subgrade distresses: Crack, Void and Interlayer Disengaging, with corresponding labels accounting for 200, 286, 173, and 23 respectively. To ensure the correctness of our built dataset, these annotations are labeled and confirmed jointly by two GPR domain experts with years of experience.  Examples of these subgrade distresses with their corresponding main view and top view images are illustrated in Fig \ref{fig:2}.

Formally, a GPR sample in the Multi-View 3D-GPR dataset is represented as a triple, denoted as $(x^M,x^T,y)$, where $x^M$, $x^T$ stand the main view image, the top view image respectively, and $y$ indicates the ground truth label of the subgrade distress. For a subgrade distress detection task,   we have a known training sample set $D_{train} = \{ (x^M_i,x^T_i,y_i) \}$, where $i \in N_{train}$, and an unknown test sample set $D_{test} = \{ (x^M_i,x^T_i) \}$, where $i \in N_{test}$. Our objective is to train a model $\mathbf{M}$ by utilizing $D_{train}$, and subsequently predict the subgrade distress types/labels of  samples in $D_{test}$.  In the experimental section, we will evaluate our proposed GPR-MVFD and various baselines using this new Multi-View 3D-GPR benchmark.

\section{Proposed Framwork}

\subsection{Overview}
\begin{figure*}[t]
    \centering
    \includegraphics[width=17cm]{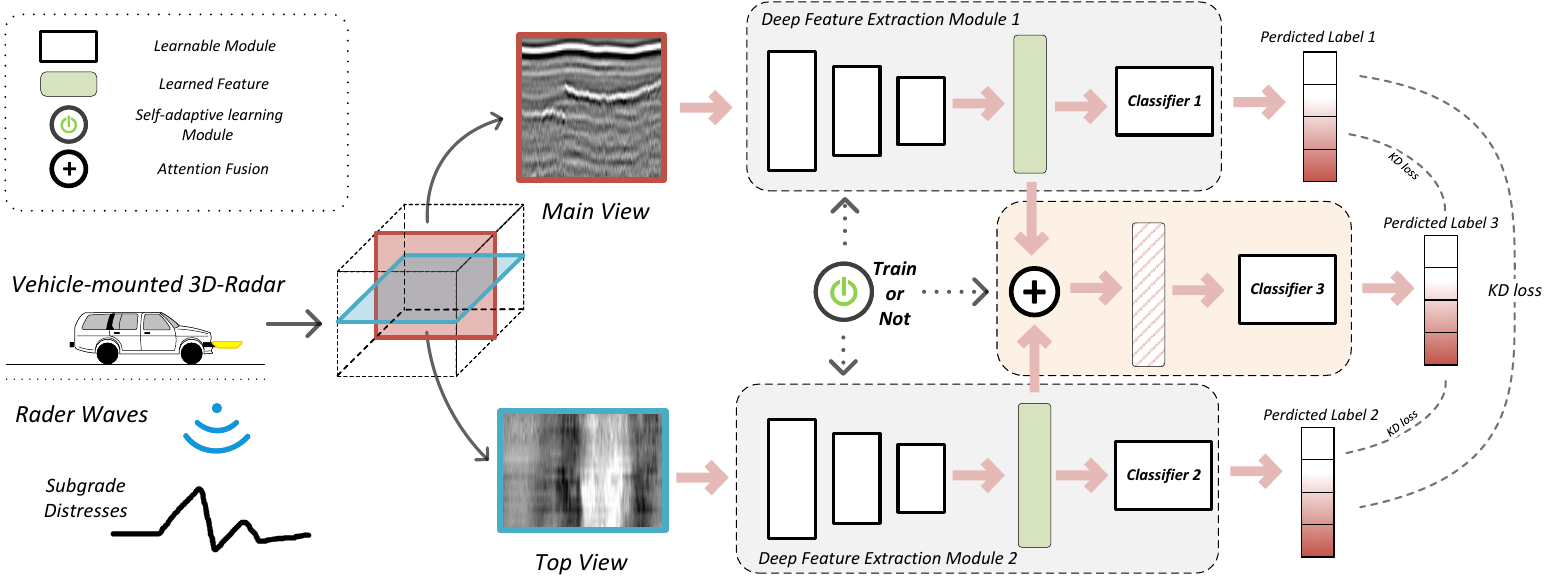}
    \caption{The overall architecture of the proposed framework GPR-MVFD, which is composed of Deep Feature Extraction Module in each branch, Multi-View
Fusion Module, Multi-View Distillation Module, and self-adaptive learning mechanism. During the inference period, the output of Multi-View Fusion Module is taken  as the final prediction.}
    \label{fig:3}
\end{figure*}

In this section, we will introduce details of  the proposed framework GPR-MVFD,  tailored for the subgrade distress detection leveraging  our multi-view GPR dataset. Figure \ref{fig:3} provides an overview of the GPR-MVFD, comprising of four key components: (1) Deep Feature Extraction Module for each branch; (2)  Multi-View Fusion Module; (3) Multi-View Distillation Module; (4) self-adaptive learning mechanism. In what follows, we will present the technical details of these components.

\subsection{Deep Feature Extraction Module}
This module utilizes Convolutional Neural Networks (CNN), such as ResNet \cite{He_2016_CVPR} or DenseNet \cite{huang2017densely}, to extract features from each view of our multi-view GPR dataset. To achieve this, we employ two CNN branches here, each corresponding to a specific view data.  Specifically, we denoted a branch as $f_1(\ )$ for feature extraction from the main view data, and another branch as $f_2(\ )$ for the top view data. Notably, these CNNs branch do not share parameters. Consequently, the extracted features from each view are represented as $h^{M} = f_1(x^{M})$ and $h^{T} =f_2(x^{T})$, where $h^{M}$ and $h^{T}$ (we omit the subscript $i$ for the simplicity here) correspond to the extracted feature from the main view and the top view respectively. These extracted features are available for direct utilization by subsequent modules.

% \begin{align}
%     H^{M} &= f_1(x^{M})\\
%     H^{T} &= f_2(x^{T})
% \end{align}

\subsection{Multi-View Fusion}
As discussed above, the main and top views of our dataset offer different perspectives, thus carrying different information. Specifically, the
main view encapsulates the travel-depth direction data, while the top view encapsulates the travel-width direction data.  Moreover, given the experts' experience knowledge, the main view and the top view may provide varying decision weights for different distresses.  For instance, void distress can be discerned by identifying the regions with highlighted feature in the main view, while the detection of the interlayer disengaging distress relies on the synergistic understanding of the hyperbolic features in the main view and the dark stain features in the top view. Additionally, previous work found that crack and void may present similar reflection patterns from a single view \cite{li2022mv}.  And the simple fusion of two views (such as summation, concatenation \cite{li2018survey}) may cause suboptimal results. Therefore, we adopt an attention-based fusion, which can automatically learn the different fusion weights for different GPR samples \cite{guo2022attention}.  This is formulated as follows:
% \begin{equation}
%   \alpha^{M} = Att(\mathbf{h^M}) \\
%   \alpha^{T} = Att(\mathbf{h^T})
% \end{equation}
\begin{align}
    \alpha^{M} &= Att(\mathbf{h^M}) \in (0,1) \\
  \alpha^{T} &= Att(\mathbf{h^T}) \, \in (0,1)\\
  \mathbf{h^F} &= \alpha^{M} \mathbf{h^M} + \alpha^{T} \mathbf{h^T} 
\end{align}
where $Att(\ )$ denotes the learnable attention mechanism shared with two views, comprising of a two-layer MLP (Multi Layer Perception) succeeded by a softmax normalization \cite{lin2017structured}. Consequently, the attention-based fusion feature $\mathbf{h^F}$ can be calculated as a weighted sum of two view features. This module allows us to obtain a robust and comprehensive fusion feature, effectively integrating information from both views, thereby enhancing the accuracy of the final decision.

\subsection{Multi-View Distillation}
Inspired by the latest advancements in knowledge distillation \cite{zhang2018deep,Guo_2020_CVPR,kim2021feature}, we integrate a multi-view distillation module into our proposed framework to further enhance the feature learning. The main idea is that our two branches corresponding to each view and the fusion module can learn reciprocally and teach each other in our framework. Though the input of these modules are different, but they all  share the same supervision signals from the distress label. Thus, these knowledges from different modules can be tranfered among them and improve each other. Initially, recalling that we have obtained three different features from a GPR sample:  $\mathbf{\mathbf{h^M}}$ from the main view, $\mathbf{h^T}$ from the top view, and $\mathbf{h^F}$ from the fusion module. 
Then, we deploy three classifiers, denoted as $C_1$, $C_2$, $C_3$ respectively, to classify $\mathbf{h^M}$, $\mathbf{h^T}$ and $\mathbf{h^F}$, as depicted in Fig \ref{fig:3}. Each classifier is constructed from a two-layer MLP followed by a softmax activation, which is used to compute the predicted distress labels distributions, denoted with  $\hat{y_1}$, $\hat{y_2}$ and $\hat{y_3}$, respectively. 
We then perform multi-view distillation via minimize the disparities among these predicted distributions \cite{hinton2015distilling}, which can be formally expressed as:
\begin{align}
    L_{KD} = D(\hat{y_1},\hat{y_2}) + D(\hat{y_2},\hat{y_1}) + D(\hat{y_3},\hat{y_1}) + D(\hat{y_3},\hat{y_2})  
\end{align}
where $L_{KD}$ is the multi-view distillation loss which we aim to minimize. The function $D(\ )$ measures the divergence between  two distributions, and we opt for the Kullback Leibler (KL) divergence with the temperature  parameter \cite{hinton2015distilling,zhang2018deep} here in our implements.

\subsection{Self-Adaptive Learning Mechanism}
As outlined previously, our objective is to further enhance each feature extraction module to learn more impactful feature from the multi-view GPR data, assisted by  the Multi-View Distillation module.
However, we empirically observed that the performance is not consistently improved  This is largely due to the fact that the existed knowledge distillation techniques most are designed under the symmetrical input, which assume identical data inputs for each module \cite{hinton2015distilling}. 
Contrarily,  the inputs of our two deep feature extraction module are different in our situation, leading to a performance gap between two branches. And  we occasionally found that this performance gap negatively impacts our distillation  process, leading to the performance degeneration. 
To overcome  the degeneration, we deploy a self-adaptive learning mechanism, which can adaptively determine whether to continue training a particular branch or not.

The mechanism mitigates the performance gap by adaptively freezing the parameters of the higher-performing branch (aka teacher) and waiting the learning of  the lower-performing branch (aka student). Consequently,  the teacher can avoid the performance degeneration, while the student still continues to train and learn from the teacher.  Following the previous work \cite{qian2022switchable}, we utilize the norm of the gradient for KL divergence loss to  quantify the performance gap $G$ in our implements, which can be  computed as $G = || y_t - y_s||_1$, where $||\ ||_1$ denotes the L-1 norm. The terms $y_t$ and $y_s$ denote the predicted distributions of any two modules, e.g. the fusion module and the main-view feature extraction module.  Till now, we also need to set a threshold $\delta  $  to  determine when to cease training. Intuitively, when $G$ remains below this threshold $\delta$  during the training period, both the teacher and student continue learning and distilling knowledge from each other.  Once $G$ exceeds $\delta$, the training of teacher is terminated and only keep the student training.  Firstly, let $y$ denote the ground true label and we can deduce the lower bound and upper bound of G as following:
\begin{align}
 || y_s - y||_1  - || y_t - y||_1 \le G < || y_s - y||_1 
\end{align}
which can further dictates  that the threshold $\delta  $ must meet:
\begin{align}\label{eq:1}
 || y_s - y||_1  - || y_t - y||_1 \le \delta < || y_s - y||_1 
\end{align}
To select an appropriate threshold according to the condition (\ref{eq:1}), we set:
\begin{align}
    \delta  &=  || y_s - y||_1  - \epsilon || y_t - y||_1   \\
    \epsilon &= e^{ - \frac{|| y_t - y||_1}{|| y_s - y||_1  + || y_t - y||_1} } \in (0,1)
\end{align}
Adopting the adaptive $\epsilon$ has several advantages: first, it ensures the condition (\ref{eq:1}) is satisfied; second, a larger $G$ will lead to a larger $\epsilon$, thus we will obtain a smaller $\delta$, and vice versa. In this way, student can catch up with the teacher. Third, it prevents the teacher from ceasing too frequently at the early stage of training \cite{qian2022switchable}.  In our implements, we empirically establish three teacher-student pairs with the self-adaptive learning mechanism, including fusion-top, fusion-main, and top-main.

\subsection{Training and Inference}
In summary, the final training loss $L$ of our framework during the training stage is formulated as:
\begin{align}
  L =  \Sigma _{i=1}^{3} CE(\hat{y_{i}},y) + L_{KD}
\end{align}
where $\hat{y_{i}}$ represents the predicted label of the classifier $C_i$ and CE(\ ) denotes the cross-entropy loss. The $L_{KD}$ stands for the multi-view distillation loss introduced in Sec 3.4.  Then the entire framework can be trained using the BP algorithm \cite{rumelhart1986learning}. 
It is worth noting that the Multi-View Fusion module and Self-Adaptive Learning Mechanism will be depreciated during the inference period which keeps our framework efficiency. Consequently, we utilize the prediction $y
_3$ from the multi-view fusion module as the final predicted result of our framework. 

\section{Experiments}
In this section, we conduct numerous experiments on our real multi-view GPR benchmark to validate the effectiveness of the proposed framework in the subgrade distress detection task.

\subsection{Baseline Methods}
% To show the effectiveness of our method in predicting multiple socioeconomic indicators, we compare our method with the following five baseline methods:

We compare the performance of our GPR-MVFD with various baselines, including the existing GPR baselines as well as the state-of-the-art multi-view, multi-modal, and knowledge distillation based methods. Unless specified otherwise, we use the DenseNet-121 \cite{huang2017densely}  as the feature extractor module in all implements of compared baselines for a fair comparison. 

\textbf{GPR baselines}.   These baslines was utilized for the GPR data in previous works, including AlexNet  \cite{xiang2019improved,krizhevsky2017imagenet}, ResNet-CBAM (CBAM for short) \cite{li2022application,woo2018cbam}. We also assess the performance of the our feature extractor module, DenseNet, individually. The classical machine learning method Support Vector Machine (SVM) \cite{xie2013gpr,hearst1998support} is also compared. It is worth noting that these methods can only leverage single view, while our dataset includes both the main and top views. Consequently, we test these baseline with each view, denoted with `-main' and `-top' respectively. 

\textbf{Multi-view based Methods}. Our proposed multi-view GPR dataset can  also  be employed with Multi-view learning methods \cite{li2018survey, yan2021deep}, including sum, max, concatenate-based Multi-view fusion. And we also compare to the recent SOTA baselines: MVCNN \cite{su2015multi}, MvFusionNet \cite{jia2019deep}, MVMSAN \cite{wang2022multi}, and MVDAN \cite{wang2022multi2}.  

\textbf{Knowledge Distillation based Methods}. The compared KD baselines include DML \cite{zhang2018deep}, KDCL \cite{Guo_2020_CVPR}, FFL \cite{kim2021feature}, SwitOKD \cite{qian2022switchable}. For these methods, each branch was regarded as a student network. 

\textbf{Multi-modal based Methods}. The recent multi-modal fusion can also be utilized  for our multi-input dataset. The compared methods include TIRG \cite{vo2019composing}  and cross-modal Transformer (transformer for short)\cite{xu2023multi}. Specially, these multi-modal fusion methods require specifying the main and auxiliary information, so we choose the main-view or top-view as the main information, denoted with `-main' and `-top' respectively.

\subsection{Experimental Settings}
All experiments were implemented on a Linux machine with a Intel(R) Xeon(R) CPU E5-2680 v4 and a NVIDIA RTX 3090 GPU. And we implement our GPR-MVFD and all baselines with Pytorch\footnote{https://pytorch.org/}.
For all experiments, the Adam optimizer \cite{kingma2014adam} was used to update parameters. For all baselines compared and GPR-MVFD, we chose hyper-parameters via the grid search across the learning rates in $\{ 1e^{-2},1e^{-3},1e^{-4} \}$, weights decay (the L-2 norm of the weights) in $\{ 1e^{-3},1e^{-4},1e^{-5} \}$ and epochs in $\{ 100,200,300 \}$.  Our labeled multi-view GPR dataset was randomly split into three disjoint parts, where $60\%$, $20\%$, $20\%$ for training, validation and testing in this subgrade distress detection benchmark. Specially, we employed the up-sampling strategy with data augmentation techniques \cite{shorten2019survey} to abbreviate the issue of imbalanced labels. 
We repeated all experiments 10 times and report the mean and standard deviation in terms of classification accuracy. Considering the practical distress detection, we need to avoid omitting. Therefore, we also report the the mean Recall and F1-score in this paper.

\subsection{Main Results}
Table \ref{tab:1} presents the results of our GPR-MVFD and the other compared baselines in the task of subgrade distress detection. First of all, our proposed GPR-MVFD achieves the state-of-the-art performance on all metrics, including the accuracy, Recall and F1-score. It supasses not only the existing GPR baselines, but also including the lasted multi-view, multi-modal, and knowledge distillation-based methods, which shows the effectiveness of our proposed methods. Second, table \ref{tab:1} also reveals that the multi-view based methods all outperform to the single-view based methods obviously. For example, the simplest multi-view learning (sum operation) demonstrates considerable improvement over existing GPR baselines. 
These results highlight the significance of multi-view information and the value of our built dataset. 
% The reason is that  our built multi-view dataset contains more spatial information, contributing the detection task. These results show the significant of multi-view information  and the advantage of our built dataset. Interestingly,   we note that that top-view based methods generally perform better than those  using main-view data, which are wildly used in GPR industry currently \cite{tong2020advances,9967436}. 
This could suggest that the spatial information of the travel-width direction provides more discrimination information for subgrade distresses detection, which reveals the potential of the top-view data for this detection task. In summary, these experiments demonstrates that the significant  improvements offered  by our proposed GPR-MVFD can be realized in detecting subgrade distresses, and our built multi-view GPR dataset can  provide more comprehensive information than the traditional single-view dataset.

\begin{table}[ht]
\caption{Subgrade distress detection results in terms of Accuracy, Recall, and F1-score.}
\setlength{\tabcolsep}{0.8mm}{

\begin{tabular}{ll |ccc }
\hline
\multicolumn{2}{c|}{\multirow{2}{*}{Methods}} & \multicolumn{3}{c}{Metrics}   \\ \cline{3-5} 
\multicolumn{2}{c|}{}  & Acc(\%) & Recall & \multicolumn{1}{c}{F1} \\ \hline
\multirow{8}{*}{GPR}        & SVM-main            &  $76.64\pm0.00$ & $0.569$ & $0.747$ \\
                                  & SVM-top             &  $74.45\pm0.00$ & $0.761$ & $0.724$ \\
                                  & AlexNet-main        &  $85.82\pm2.18$ & $0.671$ & $0.652$ \\
                                  & AlexNet-top         &  $91.90\pm0.93$  & $0.923$ & $0.927$ \\
                                  & CBAM-main    &  $83.45\pm2.18$ & $0.671$ & $0.652$ \\
                                  & CBAM-top     &  $93.52\pm1.90$ & $0.938$ & $0.936$ \\
                                  & DenseNet-main        & $86.97\pm1.91$ & $0.587$ & $0.576$ \\
                                  & DenseNet-top         & $92.23\pm0.86$ & $0.934$ & $0.934$ \\ \hline
\multirow{7}{*}{Multi-view} & sum                 & $94.31\pm1.04$ &  $0.945$  & $0.942$ \\
                                  & max                 & $94.24\pm1.87$ & $0.945$ & $0.938$ \\
                                  & concate             & $94.53\pm1.02$ & $0.948$ & $0.945$ \\
                                  & MVCNN               & $95.25\pm1.02$ & $0.952$ & $0.951$ \\
                                  & MvFusionNet         & $95.17\pm1.16$ & $0.954$ & $0.945$ \\
                                  & MVMSAN              & $94.24\pm1.67$ & $0.943$ & $0.932$ \\
                                  & MVDAN               & $93.74\pm1.87$ & $0.941$ & $0.924$  \\ \hline
\multirow{4}{*}{KD}         & DML                 & $93.95\pm1.25$ & $0.946$ & $0.942$  \\
                                  & KDCL                & $94.38\pm1.75$ & $0.937$ &  $0.940$\\
                                  & FFL                 & $95.68\pm0.64$ & $0.909$ & $0.918$ \\
                                  & SwitOKD             & $95.68\pm1.57$ & $0.909$ & $0.918$ \\ \hline
\multicolumn{1}{c}{\multirow{4}{*}{Multi-modal}} & TIRG-main & $94.53\pm1.22$  &   $0.948$  &  $0.935$                               \\
\multicolumn{1}{c}{}              & TIRG-top            & $94.53\pm1.70$ & $0.948$ &  $0.941$\\
\multicolumn{1}{c}{}              & transformer-main & $93.53\pm1.54$  &$0.937$  & $0.918$ \\
\multicolumn{1}{c}{}              & transformer-top  & $94.03\pm1.75$ & $0.943$ & $0.935$ \\ \hline
Ours                              & GPR-MVFD            & $\mathbf{96.64\pm0.95}$ & $\mathbf{0.957}$ & $\mathbf{0.961}$ \\ \hline
\end{tabular}
}
\label{tab:1}
\end{table}

\subsection{Result Analysis}
% In this section, we will further analyze our framework in more details.

\subsubsection{Ablation Study}
As introduced in Sec 3, our proposed GPR-MVFD comprises several modules and we investigate the impact of each module as demonstrated in  Table \ref{tab:2}.  In this table, case A represents the single view based method (SV for short) and we use a single DenseNet-121 here. Next, case B expands on case A by incorporating multi-view data (MV) as input. Likewise,  case C adds the multi-view fusion module (attention) and case D adds the multi-view distillation module (distillation). Finally, our GPR-MVFD can be regarded as the case D supplemented with the Self-Adaptive Learning Mechanism (AM). As the trend of accuracy in Table \ref{tab:2} shows,   each module in our framework has a positive impact on the performance, with the multi-view input and the self-adaptive learning mechanism making particularly notable contributions.

% The table illustrates the accuracy statistics with the modules adding the single view baseline.  

% \begin{figure}[h]
%     \centering
%     \includegraphics[width=8cm]{Fig/abalation.pdf}
%     \caption{Caption}
% \end{figure}

\begin{table}[ht]
\centering
\caption{Ablation study for our proposed GPR-MVFD.}
\setlength{\tabcolsep}{1.5mm}{
\begin{tabular}{c|ccccc|c|c} 
\hline
Case & SV & MV & Att & Dist & AM & Acc & $\Delta$ \\ \hline
A&\checkmark            &            &                  &              &    &    $92.23$  & - \\ 
B&\checkmark            &      \checkmark      &                  &              &    &    $94.31$ & $2.08$\\
 C& \checkmark          &       \checkmark     &        \checkmark          &              &    &    $94.56$ &  $0.25$\\
D&   \checkmark       &       \checkmark     &       \checkmark           &         \checkmark     &    & $95.10$   & $0.54$ \\
Ours  &   \checkmark      &  \checkmark          &      \checkmark            &    \checkmark          & \checkmark   &      $96.64$ &  $1.54$\\
\hline
\end{tabular}
}
\label{tab:2}
\end{table}

\subsubsection{The Effect of Data Augmentation}
  We empirically found that data augmentation techniques are vital for alleviating overfitting and enhancing the generalization capabilities of deep learning in in GPR data exploitation. Consequently, the detection performance may increase further with data augmentation techniques. This finding aligns with the conclusion in \cite{tong2020advances} that the strategic application of data augmentation methods play a crucial role in optimizing the performance of deep learning models for 3D-GPR tasks. We investigate the effects of different data augmentation techniques in the subgrade distress detection task. The combination of the resizing and random flipping  is deemed as the baseline method, commonly used in the computer vision field \cite{shorten2019survey}.  In addition to the baseline, we also assess the performance of the recent data augmentation techniques combined with the baseline using our GPR-MVFD.  Fig 4 shows all results and we find that Cutout \cite{devries2017improved}  achieves the best performance compare with others. To the best of our knowledge, we are the first to identify that the Cutout is suitable for the GPR data.

% \begin{figure} [t!]
% 	\centering
% 	\subfloat[2112e12e]{
% 		\includegraphics[width=7cm]{Fig/test.pdf}}
% 	\\
% 	\subfloat{
% 		\includegraphics[width=7cm]{Fig/time.pdf} }
% 	\caption{The MFCVs in silico under BC when the amplitudes of the simulated signals a) don't meet~(\ref{eq6}); b) meet~(\ref{eq6}). }
% 	\label{fig3} 
% \end{figure}

\begin{figure}[h]
    \centering
    \includegraphics[width=8cm]{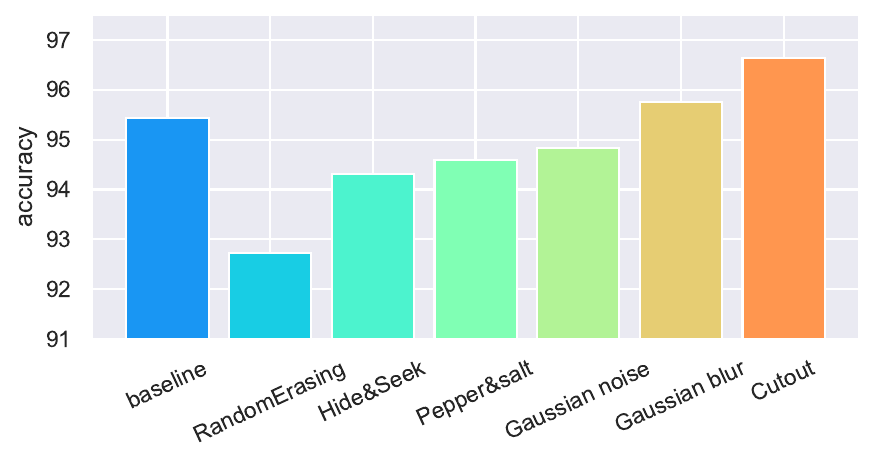}
    \caption{The performance comparisons with different data augmentations.}
\end{figure}

\subsubsection{Efficiency}
To further validate the efficiency of our proposed GPR-MVFD, we present the average inference time of a sample and the Float point Operations (FLOPs) of different methods in Fig 5.  As we can see, the inference time and FLOPs of our GPR-MVFD  still keep the same order of magnitude as the single-view based DenseNet, and significantly more efficient than the 3D-DenseNet, which is used for the 3D C-scan data. These results suggest that our GPR-MVFD does not require substantial computational resources.

\begin{figure}[h]
    \centering
    \includegraphics[width=7.5cm]{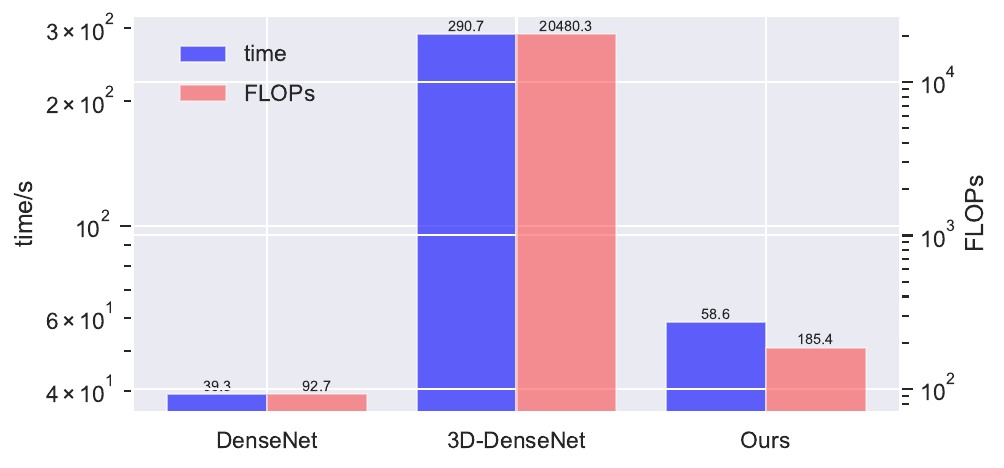}
    \caption{Running time and Float point Operations (FLOPs) comparisons.}
\end{figure}

\begin{figure}[htbp]
    \centering
    \includegraphics[width=8cm]{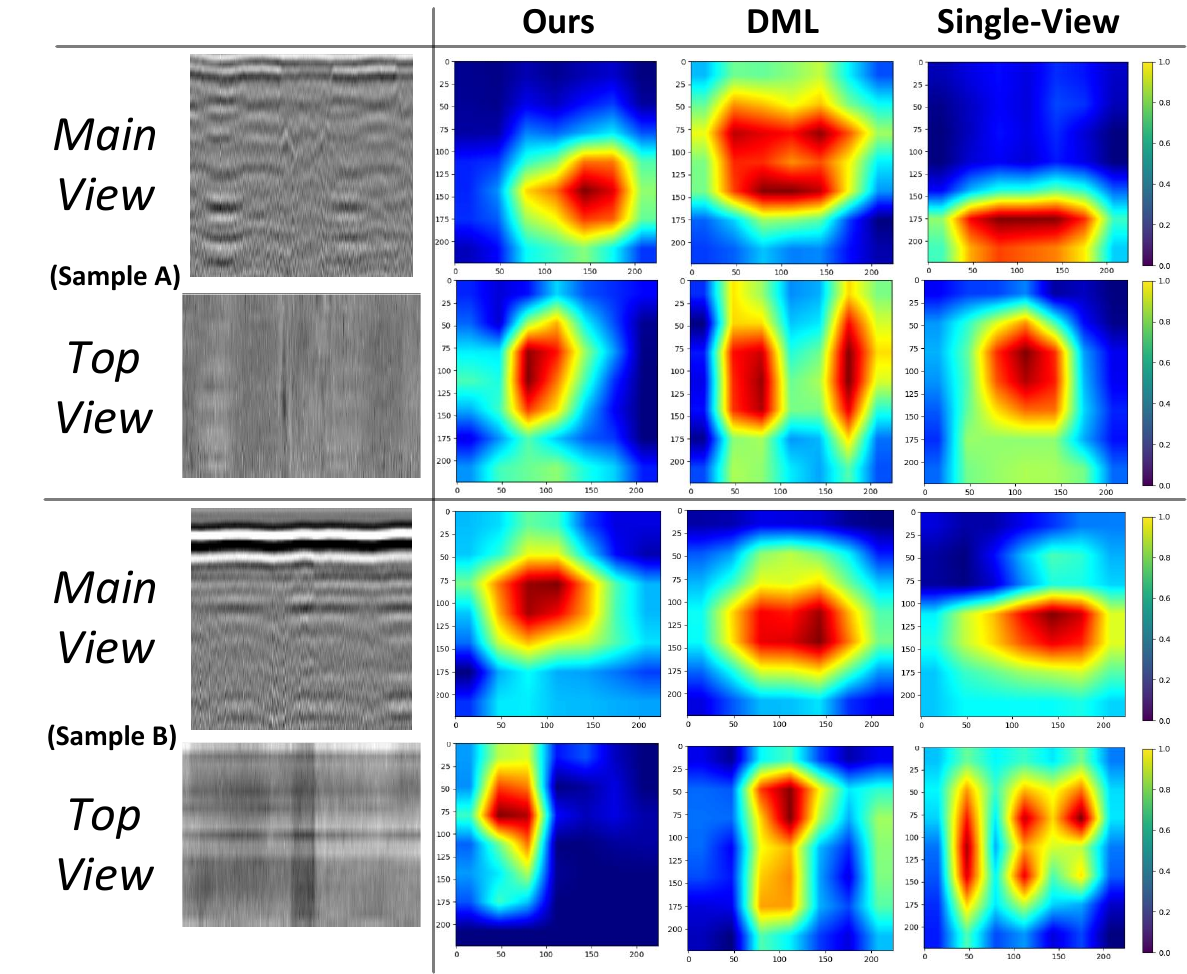}
    \caption{Explainable visualizations of our GPR-MVFD using Grad-CAM++ (better with color).}
    \label{fig:6}
\end{figure}

\subsubsection{Explainable Visualization}
Here we investigate weather our GPR-MVFD can identify significant regions from the GPR dataset. Consequently, we employ the Grad-CAM++ \cite{chattopadhay2018grad}  to produce the heatmap of our framework on the input image. We also compared our  GPR-MVFD with the single-view based DenseNet and multi-view based DML.  Fig \ref{fig:6} exhibits the output of a crack sample \textbf{A} and a void sample \textbf{B} using the Grad-CAM++.  First, we can find that GPR-MVFD focus the more significant and smaller region than the other methods, while DML and DenseNet focus overly large region of the input image resulting in suboptimal detection performance. Taking the top view of the sample \textbf{A} as a example, our GPR-MVFD  identifies  the correct  distress region in line with the expert experience, while the DML focuses the wrong region and DenseNet focuses on an excessively large irrelevant region. These visualization results validate that our GPR-MVFD can find the significant region and further demonstrates its effectiveness.

% As shown in Fig 6, the color is more close to the red, which means it get more focuses from a model. 

\section{Conclusion}
To improve the performance of the subgrade distresses detection based on 3D-GPR, we introduce a novel methodology, which utilizes the multi-view information from the original 3D-GPR data. And we also built a multi-view GPR dataset with expert-annotated labels, serving as a distress detection benchmark. This new dataset provides richer spatial information than A-scan and B-scan data, while maintaining computational efficiency compared to C-scan data.  Then, we  develop a novel multi-branch framework GPR-MVFD, specifically designed for the multi-view GPR dataset.   
 GPR-MVFD integrates knowledge distillation and attention-based fusion to facilitate significant feature extraction for subgrade distresses. We also adopt a self-adaptive learning mechanism to stabilize training. Extensive experiments conducted on the GPR benchmark demonstrate the effectiveness and efficiency of our proposed framework. We also showcases the explanatory visualizations produced by our GPR-MVFD, and  discover Cutout is suitable for the GPR data. We will release the constructed multi-view GPR dataset with expert-annotated labels and the source codes of the proposed framework. 
 We hope this work could boost the automatic detection based on 3D-GPR data, and enables more accurate and efficient subgrade distress detection in practical applications.

% \vfill\pagebreak

% \section{REFERENCES}
% \label{sec:refs}

% List and number all bibliographical references at the end of the
% paper. The references can be numbered in alphabetic order or in
% order of appearance in the document. When referring to them in
% the text, type the corresponding reference number in square
% brackets as shown at the end of this sentence \cite{C2}. An
% additional final page (the fifth page, in most cases) is
% allowed, but must contain only references to the prior
% literature.


% \vfill\pagebreak
\bibliographystyle{IEEE}
\bibliography{3DGPR}

\begin{thebibliography}{10}

\bibitem{khudoyarov2020three}
Shekhroz Khudoyarov, Namgyu Kim, and Jong-Jae Lee,
\newblock ``Three-dimensional convolutional neural network--based underground
  object classification using three-dimensional ground penetrating radar
  data,''
\newblock {\em Structural Health Monitoring}, vol. 19, no. 6, pp. 1884--1893,
  2020.

\bibitem{rasol2022gpr}
Mezgeen Rasol, Jorge~C Pais, Vega P{\'e}rez-Gracia, Mercedes Solla, Francisco~M
  Fernandes, Simona Fontul, David Ayala-Cabrera, Franziska Schmidt, and Hossein
  Assadollahi,
\newblock ``Gpr monitoring for road transport infrastructure: A systematic
  review and machine learning insights,''
\newblock {\em Construction and Building Materials}, vol. 324, pp. 126686,
  2022.

\bibitem{wang2010automatic}
Zhe~Wendy Wang, Mengchu Zhou, Gregory~G Slabaugh, Jiefu Zhai, and Tong Fang,
\newblock ``Automatic detection of bridge deck condition from ground
  penetrating radar images,''
\newblock {\em IEEE transactions on automation science and engineering}, vol.
  8, no. 3, pp. 633--640, 2010.

\bibitem{bachiri2018bridge}
Tahar Bachiri, Abdellatif Khamlichi, and Mohammed Bezzazi,
\newblock ``Bridge deck condition assessment by using gpr: a review,''
\newblock in {\em MATEC Web of Conferences}. EDP Sciences, 2018, vol. 191, p.
  00004.

\bibitem{DINH2018292}
Kien Dinh, Nenad Gucunski, and Trung~H. Duong,
\newblock ``An algorithm for automatic localization and detection of rebars
  from gpr data of concrete bridge decks,''
\newblock {\em Automation in Construction}, vol. 89, pp. 292--298, 2018.

\bibitem{li2022mv}
Nansha Li, Renbiao Wu, Haifeng Li, Huaichao Wang, Zhongcheng Gui, and Dezhen
  Song,
\newblock ``Mv-gprnet: Multi-view subsurface defect detection network for
  airport runway inspection based on gpr,''
\newblock {\em Remote Sensing}, vol. 14, no. 18, pp. 4472, 2022.

\bibitem{lecun2015deep}
Yann LeCun, Yoshua Bengio, and Geoffrey Hinton,
\newblock ``Deep learning,''
\newblock {\em nature}, vol. 521, no. 7553, pp. 436--444, 2015.

\bibitem{voulodimos2018deep}
Athanasios Voulodimos, Nikolaos Doulamis, Anastasios Doulamis, Eftychios
  Protopapadakis, et~al.,
\newblock ``Deep learning for computer vision: A brief review,''
\newblock {\em Computational intelligence and neuroscience}, vol. 2018, 2018.

\bibitem{He_2016_CVPR}
Kaiming He, Xiangyu Zhang, Shaoqing Ren, and Jian Sun,
\newblock ``Deep residual learning for image recognition,''
\newblock in {\em Proceedings of the IEEE Conference on Computer Vision and
  Pattern Recognition (CVPR)}, June 2016.

\bibitem{Lin_2017_ICCV}
Tsung-Yi Lin, Priya Goyal, Ross Girshick, Kaiming He, and Piotr Dollar,
\newblock ``Focal loss for dense object detection,''
\newblock in {\em Proceedings of the IEEE International Conference on Computer
  Vision (ICCV)}, Oct 2017.

\bibitem{He_2017_ICCV}
Kaiming He, Georgia Gkioxari, Piotr Dollar, and Ross Girshick,
\newblock ``Mask r-cnn,''
\newblock in {\em Proceedings of the IEEE International Conference on Computer
  Vision (ICCV)}, Oct 2017.

\bibitem{tong2020advances}
Zheng Tong, Jie Gao, and Dongdong Yuan,
\newblock ``Advances of deep learning applications in ground-penetrating radar:
  A survey,''
\newblock {\em Construction and Building Materials}, vol. 258, pp. 120371,
  2020.

\bibitem{7572995}
Qingxu Dou, Lijun Wei, Derek~R. Magee, and Anthony~G. Cohn,
\newblock ``Real-time hyperbola recognition and fitting in gpr data,''
\newblock {\em IEEE Transactions on Geoscience and Remote Sensing}, vol. 55,
  no. 1, pp. 51--62, 2017.

\bibitem{9702139}
Li~Liu, Hang Yu, Hang Xu, Bingjie Wang, and Jingxia Li,
\newblock ``Underground object classification using deep 3-d convolutional
  networks and multiple mirror encoding for gpr data,''
\newblock {\em IEEE Geoscience and Remote Sensing Letters}, vol. 19, pp. 1--5,
  2022.

\bibitem{giannakis2019machine}
Iraklis Giannakis, Antonios Giannopoulos, and Craig Warren,
\newblock ``A machine learning-based fast-forward solver for ground penetrating
  radar with application to full-waveform inversion,''
\newblock {\em IEEE Transactions on Geoscience and Remote Sensing}, vol. 57,
  no. 7, pp. 4417--4426, 2019.

\bibitem{tong2020pavement}
Zheng Tong, Dongdong Yuan, Jie Gao, Yongfeng Wei, and Hui Dou,
\newblock ``Pavement-distress detection using ground-penetrating radar and
  network in networks,''
\newblock {\em Construction and Building Materials}, vol. 233, pp. 117352,
  2020.

\bibitem{wang2019human}
Wei Wang,
\newblock ``Human detection based on radar sensor network in natural
  disaster,''
\newblock {\em Geological Disaster Monitoring Based on Sensor Networks}, pp.
  109--134, 2019.

\bibitem{rs13122375}
Juncai Xu, Jingkui Zhang, and Weigang Sun,
\newblock ``Recognition of the typical distress in concrete pavement based on
  gpr and 1d-cnn,''
\newblock {\em Remote Sensing}, vol. 13, no. 12, 2021.

\bibitem{TONG201869}
Zheng Tong, Jie Gao, and Haitao Zhang,
\newblock ``Innovative method for recognizing subgrade defects based on a
  convolutional neural network,''
\newblock {\em Construction and Building Materials}, vol. 169, pp. 69--82,
  2018.

\bibitem{su2023end}
Yang Su, Jun Wang, Danqi Li, Xiangyu Wang, Lei Hu, Yuan Yao, and Yuanxin Kang,
\newblock ``End-to-end deep learning model for underground utilities
  localization using gpr,''
\newblock {\em Automation in Construction}, vol. 149, pp. 104776, 2023.

\bibitem{kim2021novel}
Namgyu Kim, Sehoon Kim, Yun-Kyu An, and Jong-Jae Lee,
\newblock ``A novel 3d gpr image arrangement for deep learning-based
  underground object classification,''
\newblock {\em International Journal of Pavement Engineering}, vol. 22, no. 6,
  pp. 740--751, 2021.

\bibitem{rs14071593}
Jianyu Ling, Rongyi Qian, Ke~Shang, Linyan Guo, Yu~Zhao, and Dongyi Liu,
\newblock ``Research on the dynamic monitoring technology of road subgrades
  with time-lapse full-coverage 3d ground penetrating radar (gpr),''
\newblock {\em Remote Sensing}, vol. 14, no. 7, 2022.

\bibitem{hinton2015distilling}
Geoffrey Hinton, Oriol Vinyals, and Jeff Dean,
\newblock ``Distilling the knowledge in a neural network,''
\newblock {\em arXiv preprint arXiv:1503.02531}, 2015.

\bibitem{zhang2018deep}
Ying Zhang, Tao Xiang, Timothy~M Hospedales, and Huchuan Lu,
\newblock ``Deep mutual learning,''
\newblock in {\em Proceedings of the IEEE conference on computer vision and
  pattern recognition}, 2018, pp. 4320--4328.

\bibitem{qian2022switchable}
Biao Qian, Yang Wang, Hongzhi Yin, Richang Hong, and Meng Wang,
\newblock ``Switchable online knowledge distillation,''
\newblock in {\em Computer Vision--ECCV 2022: 17th European Conference, Tel
  Aviv, Israel, October 23--27, 2022, Proceedings, Part XI}. Springer, 2022,
  pp. 449--466.

\bibitem{huang2017densely}
Gao Huang, Zhuang Liu, Laurens Van Der~Maaten, and Kilian~Q Weinberger,
\newblock ``Densely connected convolutional networks,''
\newblock in {\em Proceedings of the IEEE conference on computer vision and
  pattern recognition}, 2017, pp. 4700--4708.

\bibitem{li2018survey}
Yingming Li, Ming Yang, and Zhongfei Zhang,
\newblock ``A survey of multi-view representation learning,''
\newblock {\em IEEE transactions on knowledge and data engineering}, vol. 31,
  no. 10, pp. 1863--1883, 2018.

\bibitem{guo2022attention}
Meng-Hao Guo, Tian-Xing Xu, Jiang-Jiang Liu, Zheng-Ning Liu, Peng-Tao Jiang,
  Tai-Jiang Mu, Song-Hai Zhang, Ralph~R Martin, Ming-Ming Cheng, and Shi-Min
  Hu,
\newblock ``Attention mechanisms in computer vision: A survey,''
\newblock {\em Computational Visual Media}, vol. 8, no. 3, pp. 331--368, 2022.

\bibitem{lin2017structured}
Zhouhan Lin, Minwei Feng, Cicero Nogueira~dos Santos, Mo~Yu, Bing Xiang, Bowen
  Zhou, and Yoshua Bengio,
\newblock ``A structured self-attentive sentence embedding,''
\newblock {\em arXiv preprint arXiv:1703.03130}, 2017.

\bibitem{Guo_2020_CVPR}
Qiushan Guo, Xinjiang Wang, Yichao Wu, Zhipeng Yu, Ding Liang, Xiaolin Hu, and
  Ping Luo,
\newblock ``Online knowledge distillation via collaborative learning,''
\newblock in {\em Proceedings of the IEEE/CVF Conference on Computer Vision and
  Pattern Recognition (CVPR)}, June 2020.

\bibitem{kim2021feature}
Jangho Kim, Minsung Hyun, Inseop Chung, and Nojun Kwak,
\newblock ``Feature fusion for online mutual knowledge distillation,''
\newblock in {\em 2020 25th International Conference on Pattern Recognition
  (ICPR)}. IEEE, 2021, pp. 4619--4625.

\bibitem{rumelhart1986learning}
David~E Rumelhart, Geoffrey~E Hinton, and Ronald~J Williams,
\newblock ``Learning representations by back-propagating errors,''
\newblock {\em nature}, vol. 323, no. 6088, pp. 533--536, 1986.

\bibitem{xiang2019improved}
Zhongming Xiang, Abbas Rashidi, and Ge~Ou,
\newblock ``An improved convolutional neural network system for automatically
  detecting rebar in gpr data,''
\newblock in {\em Computing in Civil Engineering 2019: Data, Sensing, and
  Analytics}, pp. 422--429. American Society of Civil Engineers Reston, VA,
  2019.

\bibitem{krizhevsky2017imagenet}
Alex Krizhevsky, Ilya Sutskever, and Geoffrey~E Hinton,
\newblock ``Imagenet classification with deep convolutional neural networks,''
\newblock {\em Communications of the ACM}, vol. 60, no. 6, pp. 84--90, 2017.

\bibitem{li2022application}
Jiadai Li, Ding Yang, Cheng Guo, Chenggao Ji, Yangchao Jin, Haijiao Sun, and
  Qing Zhao,
\newblock ``Application of gpr system with convolutional neural network
  algorithm based on attention mechanism to oil pipeline leakage detection,''
\newblock {\em New Advances in Geology and Engineering Technology of
  Unconventional Oil and Gas}, vol. 10, pp. 863730, 2022.

\bibitem{woo2018cbam}
Sanghyun Woo, Jongchan Park, Joon-Young Lee, and In~So Kweon,
\newblock ``Cbam: Convolutional block attention module,''
\newblock in {\em Proceedings of the European conference on computer vision
  (ECCV)}, 2018, pp. 3--19.

\bibitem{xie2013gpr}
Xiongyao Xie, Pan Li, Hui Qin, Lanbo Liu, and David~C Nobes,
\newblock ``Gpr identification of voids inside concrete based on the support
  vector machine algorithm,''
\newblock {\em Journal of Geophysics and Engineering}, vol. 10, no. 3, pp.
  034002, 2013.

\bibitem{hearst1998support}
Marti~A. Hearst, Susan~T Dumais, Edgar Osuna, John Platt, and Bernhard
  Scholkopf,
\newblock ``Support vector machines,''
\newblock {\em IEEE Intelligent Systems and their applications}, vol. 13, no.
  4, pp. 18--28, 1998.

\bibitem{yan2021deep}
Xiaoqiang Yan, Shizhe Hu, Yiqiao Mao, Yangdong Ye, and Hui Yu,
\newblock ``Deep multi-view learning methods: A review,''
\newblock {\em Neurocomputing}, vol. 448, pp. 106--129, 2021.

\bibitem{su2015multi}
Hang Su, Subhransu Maji, Evangelos Kalogerakis, and Erik Learned-Miller,
\newblock ``Multi-view convolutional neural networks for 3d shape
  recognition,''
\newblock in {\em Proceedings of the IEEE international conference on computer
  vision}, 2015, pp. 945--953.

\bibitem{jia2019deep}
Kui Jia, Jiehong Lin, Mingkui Tan, and Dacheng Tao,
\newblock ``Deep multi-view learning using neuron-wise correlation-maximizing
  regularizers,''
\newblock {\em IEEE Transactions on Image Processing}, vol. 28, no. 10, pp.
  5121--5134, 2019.

\bibitem{wang2022multi}
Wenju Wang, Xiaolin Wang, Gang Chen, and Haoran Zhou,
\newblock ``Multi-view softpool attention convolutional networks for 3d model
  classification,''
\newblock {\em Frontiers in Neurorobotics}, p. 255, 2022.

\bibitem{wang2022multi2}
Wenju Wang, Yu~Cai, and Tao Wang,
\newblock ``Multi-view dual attention network for 3d object recognition,''
\newblock {\em Neural Computing and Applications}, vol. 34, no. 4, pp.
  3201--3212, 2022.

\bibitem{vo2019composing}
Nam Vo, Lu~Jiang, Chen Sun, Kevin Murphy, Li-Jia Li, Li~Fei-Fei, and James
  Hays,
\newblock ``Composing text and image for image retrieval-an empirical
  odyssey,''
\newblock in {\em Proceedings of the IEEE/CVF conference on computer vision and
  pattern recognition}, 2019, pp. 6439--6448.

\bibitem{xu2023multi}
Yahui Xu, Yi~Bin, Jiwei Wei, Yang Yang, Guoqing Wang, and Heng~Tao Shen,
\newblock ``Multi-modal transformer with global-local alignment for composed
  query image retrieval,''
\newblock {\em IEEE Transactions on Multimedia}, 2023.

\bibitem{kingma2014adam}
Diederik~P Kingma and Jimmy Ba,
\newblock ``Adam: A method for stochastic optimization,''
\newblock {\em arXiv preprint arXiv:1412.6980}, 2014.

\bibitem{shorten2019survey}
Connor Shorten and Taghi~M Khoshgoftaar,
\newblock ``A survey on image data augmentation for deep learning,''
\newblock {\em Journal of big data}, vol. 6, no. 1, pp. 1--48, 2019.

\bibitem{devries2017improved}
Terrance DeVries and Graham~W Taylor,
\newblock ``Improved regularization of convolutional neural networks with
  cutout,''
\newblock {\em arXiv preprint arXiv:1708.04552}, 2017.

\bibitem{chattopadhay2018grad}
Aditya Chattopadhay, Anirban Sarkar, Prantik Howlader, and Vineeth~N
  Balasubramanian,
\newblock ``Grad-cam++: Generalized gradient-based visual explanations for deep
  convolutional networks,''
\newblock in {\em 2018 IEEE winter conference on applications of computer
  vision (WACV)}. IEEE, 2018, pp. 839--847.

\end{thebibliography}
% \bibliography{strings,refs}

\end{document}